\documentclass{article}

\usepackage[preprint]{tackling_climate_workshop_style}

\usepackage[utf8]{inputenc} 
\usepackage[T1]{fontenc}    
\usepackage{hyperref}       
\usepackage{url}            
\usepackage{booktabs}       
\usepackage{amsfonts}       
\usepackage{nicefrac}       
\usepackage{microtype}      

\usepackage{graphicx}

\title{Downscaling climate projections to 1 km with~single-image super resolution}

\author{%
  Petr Košťál\\
  Faculty of Information Technology\\
  Czech Technical University in Prague\\
  Thákurova 9, 160 00 Prague 6, Czechia\\
  \texttt{kostape4@fit.cvut.cz}\\
  \And
  Pavel Kordík\\
  Faculty of Information Technology\\
  Czech Technical University in Prague\\
  Thákurova 9, 160 00 Prague 6, Czechia\\
  \texttt{pavel.kordik@fit.cvut.cz}\\
  \And
  Ondřej Podsztavek\\
  Faculty of Information Technology\\
  Czech Technical University in Prague\\
  Thákurova 9, 160 00 Prague 6, Czechia\\
  \texttt{ondrej.podsztavek@fit.cvut.cz}\\
}

\begin{document}

\maketitle

\begin{abstract}
High-resolution climate projections are essential for local decision-making.
However, available climate projections have low spatial resolution (e.g.~12.5\,km), which limits their usability.
We address this limitation by leveraging single-image super-resolution models to statistically downscale climate projections to 1-km resolution.
Since high-resolution climate projections are unavailable, we train models on a~high-resolution observational gridded data set and apply them to low-resolution climate projections.
We cannot evaluate downscaled climate projections with common metrics (e.g.~pixel-wise root-mean-square error) because we lack ground-truth high-resolution climate projections.
Therefore, we evaluate climate indicators computed at weather station locations.
Experiments on daily mean temperature demonstrate that single-image super-resolution models can downscale climate projections without increasing the error of climate indicators compared to low-resolution climate projections.
\end{abstract}

\section{Introduction}

Climate projections (i.e.~simulations of climate typically until 2100) with high spatial resolution are necessary for local decision-making, such as urban planning (e.g.~assessment of urban heat islands or flood risks) or forest management (e.g.~evaluation of suitable tree species for a~specific location).
However, only climate projections with low spatial resolutions are usually available, which are often insufficient.
For example, climate projections from the European Coordinated Regional Downscaling Experiment (EURO-CORDEX) have a~resolution of 50 or 12.5\,km \citep{jacob2014}.
This resolution is obviously insufficient to resolve urban structures etc.
Therefore, our goal is to increase the spatial resolution of the climate projections to 1\,km and higher.

The task of increasing the spatial resolution of climate data is called downscaling.
There are two types of downscaling: dynamical and statistical (also known as empirical).
Dynamical downscaling relies on physics-based climate models to model climate processes in high resolution, producing physically consistent climate data, but it is computationally demanding.
Statistical downscaling establishes statistical relationships between low- and high-resolution climate data.
This type of downscaling incorporates machine learning models.
Training machine learning models is computationally demanding, but inference is fast.
This is not the only reason machine-learning models are considered for statistical downscaling of climate data \citep[see][]{rampal2024}.

There have been many attempts to downscale climate reanalysis or observations using machine learning models \citep[e.g.][]{vandal2017, price2022, oyama2023, yang2023, perez2024, sinha2025, ren2025}.
The models include variants of the super-resolution convolutional neural network (SRCNN) \citep{dong2016}, enhanced deep super-resolution network (EDSR) \citep{lim2017}, generative adversarial net (GAN) \citep{goodfellow2014}, Fourier neural operator (FNO) \citep{li2021}, and Transformer-based image-restoration model SwinIR \citep{liang2021}, based on the Shifted windows (Swin) Transformer \citep{liu2021}.

There have been fewer attempts to downscale climate projections using machine learning and evaluate them \citep{bano-medina2022, quesada-chacon2023, gonzalez-abad2023, soares2024, prasad2024}.
We can train a~model on pairs of low- and high-resolution climate projections to downscale climate projections.
However, we depend on the availability of high-resolution climate projections and are limited by their spatial resolution.
We only have observations if we want to downscale to resolutions with no climate projections available.
But here we face the challenge that we cannot create a~data set for machine learning.
We cannot pair climate projections with observations in time because there is almost no time correspondence between climate projections and observations, since climate projections are produced by free-running climate models that diverge from observations.
Therefore, we must train machine learning models on observations and transfer them to climate projections, which is potentially problematic due to distribution discrepancies.
Moreover, we cannot compute common metrics, e.g.~pixel-wise root-mean-square error (RMSE), because we do not have ground-truth high-resolution climate projections.
For example, \citet{prasad2024} evaluate on pairs of low- and high-resolution climate projections, so they are limited by their high resolution (i.e.~$0.25^\circ$).

We contribute to the statistical downscaling of climate projections to the 1-km resolution (or even higher if we have appropriate observations) with
1.~advanced single-image super-resolution models (FNO and SwinIR)
and 2.~a way to evaluate them based on climate indicators and observations from weather stations.

\section{Single-image super resolution using observational gridded data sets}

We approach statistical downscaling with single-image super resolution that treats climate data as low-resolution images and applies machine learning models to produce their high-resolution versions.
We want to downscale to a~spatial resolution with no climate projections, so we do not have pairs of low- and high-resolution climate projections for training.
Therefore, inspired by \citet{quesada-chacon2023}, we train models using a~high-resolution observational gridded data set.
Specifically, we train models with pairs consisting of
1.~a target high-resolution observational gridded sample and
2.~an input upscaled sample from the high-resolution sample.
We match the resolution of the upscaled sample and the climate projection we want to downscale.
Then, by applying the models to the climate projection, we can downscale it to a~high resolution.

Following \citet{prasad2024, ren2025} and \citet{sinha2025}, we selected a~set of single-image super-resolution models that have demonstrated strong performance in downscaling:
EDSR \citep{lim2017} from the family of convolutional network models;
FNO \citep{li2021}, a~model designed to learn solutions to partial differential equations; and a~SwinIR model
based on Transformers \citep{liang2021} that consistently beats other models.

\section{Evaluation of downscaled climate projections with climate indicators}

The evaluation of models for downscaling climate projections faces the challenge of lacking ground-truth high-resolution climate projections as introduced in Introduction.
Therefore, we evaluate climate indicators that capture trends and variability at weather station locations.
Inspired by \citet{gleckler2008}, we compute the root-mean-square error (RMSE) of a~climate indicator as:
$$\sqrt{\frac{1}{NT} \sum_{n = 1}^N \sum_{t = 1}^{T} (\hat{I}_{n, t} - I_{n, t})^2}\,,$$
where $n$ indexes $N$ weather stations,
$t$ indexes $T$ time periods (e.g.~years for which we compute the climate indicator),
$\hat{I}_{n, t}$ is the value of the climate indicator computed from the pixel of a~low-resolution or downscaled climate projection that matches the location of the weather station $n$,
and $I_{n, t}$ is its value computed from observations from the weather station $n$ (i.e.~ground truth).

\section{Experiments}

We experiment with the single-image super-resolution downscaling in a~region covering parts of Germany, Czechia, and Poland.
However, both the single-image super-resolution downscaling and evaluation with climate indicators are generalisable to other regions, given an observational gridded data set with sufficient resolution and observations from weather stations.
The code underlying the experiments is available on GitHub, at \url{https://github.com/k0stal/sisr-downscaling/}.

\subsection{Data}

As the observational gridded data set, we selected the data set from the Regional Climate Information System for Saxony, Saxony-Anhalt, and Thuringia (ReKIS)\footnote{\url{https://rekisviewer.hydro.tu-dresden.de/viewer/rekis_domain/KlimRefDS_v3.1_1961-2023.html}} that was also used by \citet{quesada-chacon2023}.
The data set was created from observations of weather stations.
We used 403 of them for evaluation (see Figure~\ref{fig:stations} in the appendix~\ref{app:evaluation}).
It is in 1\,km resolution, covers 1961--2023 and contains climate variables related to pressure, evaporation, wind, humidity, radiation, precipitation, and temperature.
For now, we experiment with the daily mean temperature.

As the climate projection to be downscaled, we selected the evaluation run (i.e.~climate projection for the past with a~boundary condition provided by a~reanalysis) of the regional climate model REMO2015 from the EURO-CORDEX.
The climate projection is at 0.11$^{\circ}$ (about 12.5\,km) resolution and covers the period 1979--2012.

\subsection{Data preparation and hyperparameters}

Selected models require that the target resolution is an integer multiple of the input resolution.
Therefore, the observational gridded data set was cropped to $400 \times 400$ pixels and upscaled to $40 \times 40$ pixels, using the \texttt{cubic\_spline} resampling in the \texttt{rioxarray} Python library.

Since our evaluation is based on trends and variability, we partitioned the observational gridded data set into training (1961--1992), validation (1993--2002) and testing sets (2003--2012) by year.
Data sets were standardised to stabilise the training.
The hyperparameters of the models were selected based on the pixel-wise RMSE computed on the validation set.
See the appendix~\ref{app:training} for details.

\subsection{Evaluation}

Following \citet{vautard2021}, we selected the following climate indicators that capture both the trend and variability of the daily mean temperature:
1.~annual mean of daily mean temperature (TG) in Kelvins,
2.~annual growing degree days (GDD) that inform about heat accumulation relevant for plant growth,
3.~annual cooling degree days (CDD) that characterize the energy demand for cooling, and
4.~annual heating degree days (HDD) that assess the energy demand for heating.
The GDD, CDD, and HDD are the \emph{cumulative degree days} for days when the daily mean temperature is above 5, above 22, and below 15.5\,$^{\circ}$C, respectively.
These three are in Kelvin days.
See the appendix~\ref{app:evaluation} for details.

\subsection{Results}

\begin{figure}
  \centering
  \includegraphics[
    alt={Low-resolution and downscaled climate projection for May 1, 2003},
    width=\linewidth
  ]{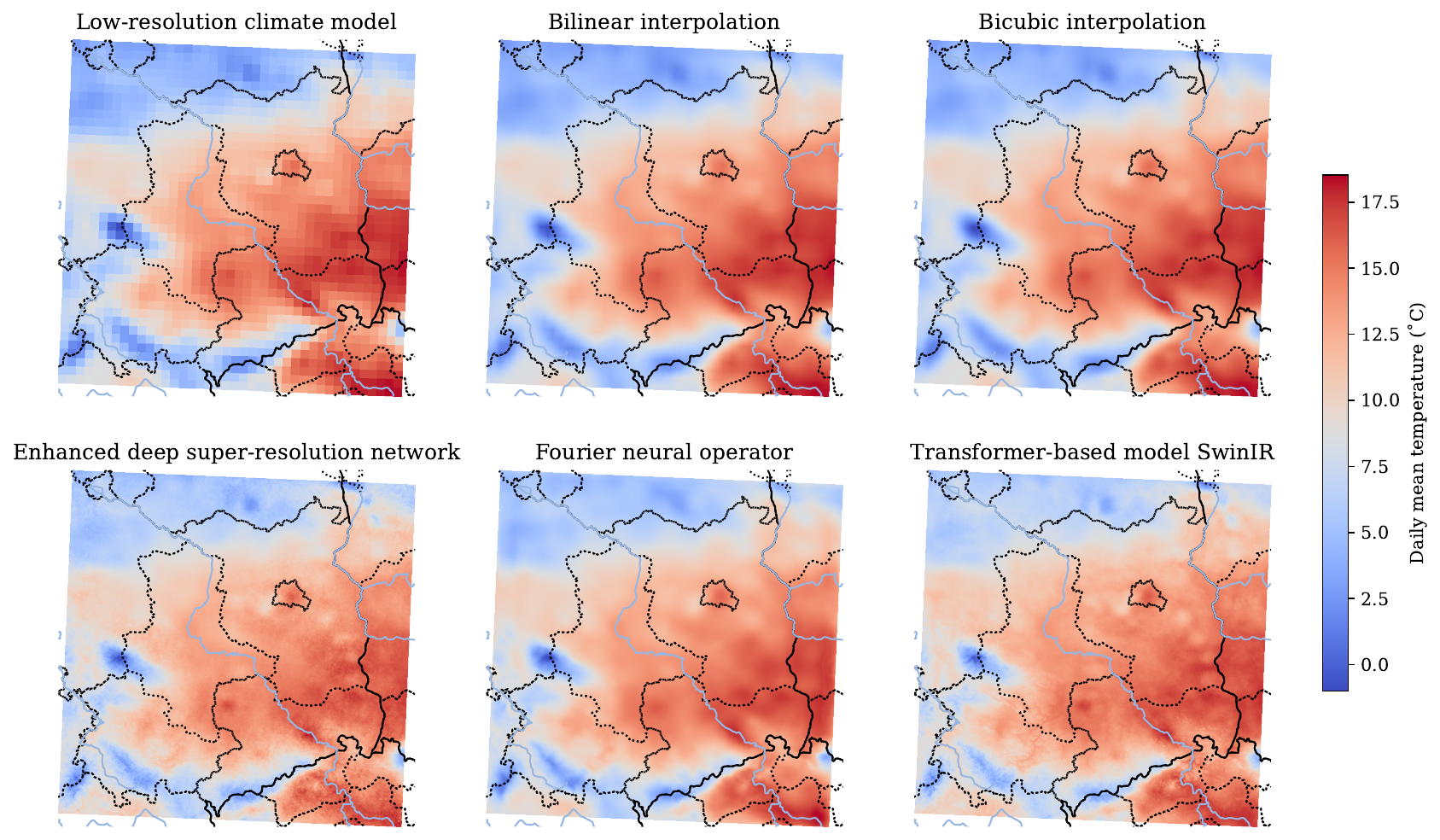}
  \caption{Low-resolution (upper left) and downscaled climate projection for May 1, 2003}
  \label{fig:sample}
\end{figure}

Table~\ref{metrics} shows the RMSE of the climate indicators computed for the downscaled climate projection during the test period (i.e.~2003--2012) with different methods.
We used bilinear and bicubic interpolation and the low-resolution regional climate model REMO2015 as baselines.
Single-image super-resolution models outperform baselines on most of the climate indicators.
An exception is CDD, where the differences are negligible.
Most importantly, all models surpass REMO2015.
Figure~\ref{fig:sample} shows a~sample low-resolution climate projection of REMO2015 for May 1, 2003, and its downscaled versions.

\begin{table}
  \caption{RMSE of climate indicators on the test period (2003--2012)}
  \label{metrics}
  \centering
  \begin{tabular}{lrrrrr}
    \toprule
    Method & TG & GDD & CDD & HDD \\
    \midrule
    REMO2015 & 0.88 & 258.58 & 31.30 & 261.00 \\
    Bilinear & 0.88 & 256.77 & 30.23 & 265.38 \\
    Bicubic  & 0.87 & 253.69 & 30.34 & 260.30 \\
    EDSR     & 0.78 & 241.56 & 30.84 & 233.54 \\
    FNO      & 0.82 & 250.92 & 30.80 & 243.84 \\
    SwinIR   & 0.77 & 241.30 & 31.02 & 227.90 \\
    \bottomrule
  \end{tabular}
\end{table}

\section{Conclusion}

We presented single-image super-resolution models for downscaling climate projections to 1-km resolution and a~way to evaluate their performance in downscaling climate projections with climate indicators.
The results show that these models can downscale climate projections without increasing the error compared to low-resolution ones.

This is research in progress, so many problems remain for the future.
In Figure~\ref{fig:sample}, climate projections downscaled with single-image super-resolution models show much more local structure.
Although our way of evaluation shows that single-image super-resolution models are better, it does not capture the added value of the local structure.
Therefore, we have to extend it to ways to assess spatial properties.
Furthermore, we need to implement probabilistic models as uncertainty quantification is important to tackle climate change.
Exploring other climate variables, such as daily minimum and maximum temperatures and precipitation, is also necessary.
Moreover, incorporating auxiliary data (e.g.~elevation) can enhance performance, particularly in regions with complex terrain.
Finally, we did not include diffusion models, which are gaining increasing attention \citep[e.g.][]{mardani2025}.

\begin{ack}
This work was supported by the project Efficient Exploration of Climate Data Locally (FOCAL) funded by the European Union (CINEA), grant No.~101137787, and Student Summer Research Program 2025 of the FIT CTU in Prague.
\end{ack}

\bibliographystyle{abbrvnat}
\bibliography{bibliography}

\newpage
\appendix

\section{Training details}
\label{app:training}

In the following, we summarise the hyperparameter search and training configurations for each model.
We trained the models for 350 epochs using a cosine annealing learning rate scheduler and a~batch size of 32.

The EDSR was trained with the $L^1$ loss and the Adam algorithm (a learning rate of $1 \times 10^{-4}$ and a~weight decay of $1 \times 10^{-5}$).
We conducted a~hyperparameter search over network widths \{64, 128, 256\} and depths \{16, 32, 64\}.
The best configuration was a~width of 128 and a~depth of 64 layers.

The FNO was trained with the mean-square error loss and the Adam algorithm (a learning rate of $1 \times 10^{-3}$).
We conducted a~hyperparameter search over the number of layers \{1, 4, 7\}, hidden channel sizes \{16, 32, 64\}, and Fourier modes \{8, 12, 14\}.
The best configuration was a~7-layer architecture with 64 hidden channels and 14 Fourier modes in both spatial dimensions.

The SwinIR was trained with the $L^1$ loss and the Adam with decoupled weight decay (AdamW) algorithm (a learning rate of $1 \times 10^{-4}$ and a~weight decay of $1 \times 10^{-5}$).
We conducted a~hyperparameter search over window sizes \{4, 5, 8\}, layers per block \{4, 6, 8\}, and the number of residual Swin transformer blocks (RSTBs) \{4, 6, 8\}.
The best configuration was 6 RSTBs, each containing 6 layers, with a~hidden dimension of 180, a~window size of 8, and 6 attention heads per layer.

In Table~\ref{validation-metrics}, we provide the pixel-wise RMSE, mean peak signal-to-noise ratio (PSNR) and mean structural similarity index measure (SSIM) computed on the ReKIS observational gridded \emph{validation} set.
We provide the PSNR and SSIM for completeness because they are used to evaluate single-image super-resolution models.
However, they are more relevant to images \citep{wang2024} than to climate data.

\begin{table}
  \caption{Pixel-wise RMSE, PSNR, SSIM on the ReKIS observational gridded \emph{validation} set}
  \label{validation-metrics}
  \centering
  \begin{tabular}{lrrr}
    \toprule
    Method & pixel-wise RMSE (K) & PSNR (dB) & SSIM \\
    \midrule
    Bilinear & 0.275 & 45.3 & 0.940 \\
    Bicubic  & 0.260 & 45.8 & 0.942 \\
    EDSR     & 0.037 & 63.3 & 0.998 \\
    FNO      & 0.052 & 59.7 & 0.996 \\
    SwinIR   & 0.032 & 64.7 & 0.998 \\
    \bottomrule
  \end{tabular}
\end{table}

\section{Evaluation details}
\label{app:evaluation}

We implemented climate indicators with the \texttt{xclim} Python library \citep{bourgault2023}.

\begin{figure}
  \centering
  \includegraphics[
    alt={ReKIS evaluation weather stations},
    width=.4\linewidth
  ]{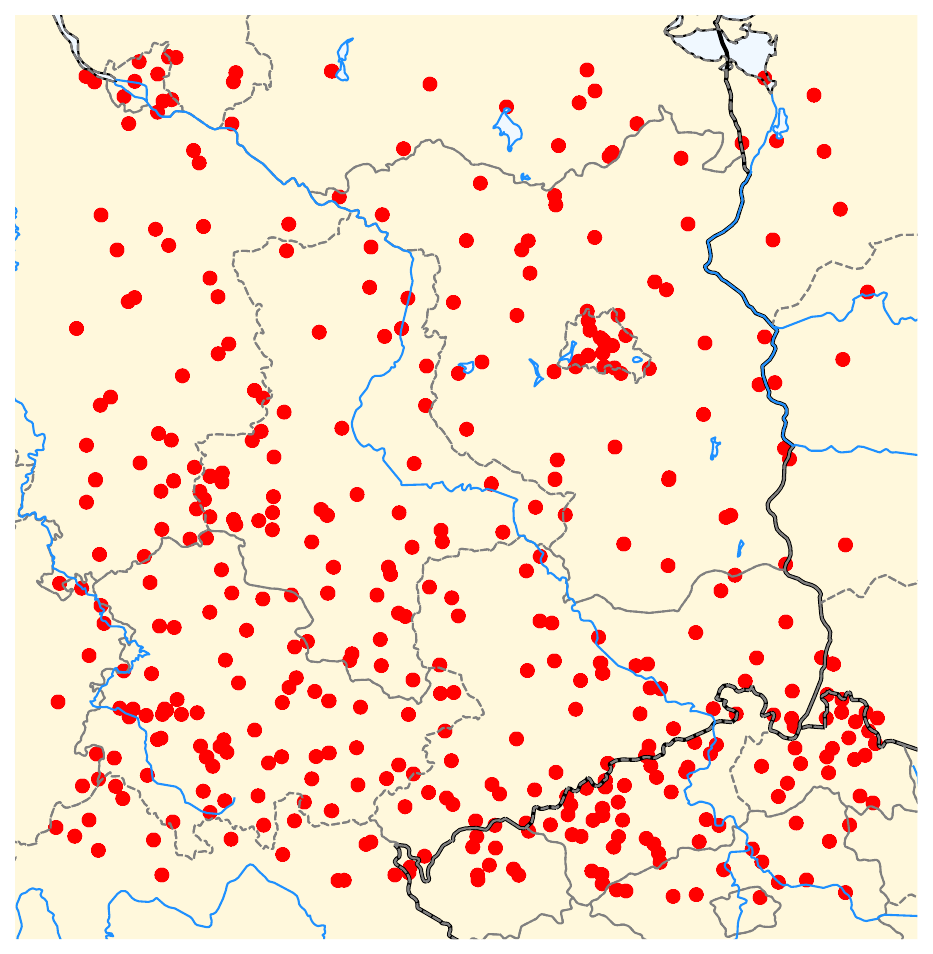}
  \caption{ReKIS evaluation weather stations}
  \label{fig:stations}
\end{figure}

Figure~\ref{fig:stations} shows the locations of the 403 ReKIS evaluation weather stations.

\end{document}